\documentclass[a4paper,11pt,DIV=14,abstracton,headings=small,twocolumn]{scrartcl}

\usepackage{todonotes} 
\usepackage{hyperref}
\usepackage{natbib}

\newcommand{\URL}[1]{\href{http://#1}{\texttt{#1}}}

\begin{document}
\title{RoboCupSimData: A RoboCup soccer research dataset} 
\author{Olivia Michael$^1$, Oliver Obst$^{*1}$, \\
Falk Schmidsberger$^2$, Frieder Stolzenburg$^2$}
\date{}
\publishers{\small $^1$: Western Sydney University, Centre for Research in Mathematics\\
$^2$: Harz University of Applied Sciences, Automation and Computer Sciences Department\\
$^*$: Corresponding author: \href{mailto:o.obst@westernsydney.edu.au}{\texttt{o.obst@westernsydney.edu.au}}
}

\twocolumn[
	\begin{@twocolumnfalse}
	\maketitle
	\vspace*{-1cm}		
		\begin{abstract}
	RoboCup is an international scientific robot competition in which teams of
	multiple robots compete against each other. Its different leagues provide many
	sources of robotics data, that can be used for further analysis and application
	of machine learning. This paper describes a large dataset from games of some of
	the top teams (from 2016 and 2017) in RoboCup Soccer Simulation League (2D),
	where teams of 11 robots (agents) compete against each other. Overall, we used
	10 different teams to play each other, resulting in 45 unique pairings. For each
	pairing, we ran 25~matches (of 10\,mins), leading to 1125~matches or more than
	180 hours of game play. The generated CSV files are 17\,GB of data (zipped), or
	229\,GB (unzipped). The dataset is unique in the sense that it contains both the
	ground truth data (global, complete, noise-free information of all objects on
	the field), as well as the noisy, local and incomplete percepts of each robot.
	These data are made available as CSV files, as well as in the original soccer
	simulator formats.\\
		\end{abstract}
	\end{@twocolumnfalse}
]

\section{Introduction}
 
RoboCup is an international scientific robot competition in which teams of
multiple robots compete against each other. The RoboCup soccer leagues provide
platforms for a number of challenges in robotics research, including locomotion,
vision, real-time decision making, dealing with partial information, multi-robot
coordination and teamwork. In RoboCup, several different leagues exist to
emphasize specific research problems by using different kinds of robots and
rules.
There are different soccer leagues in the RoboCup with different types and sizes
of hardware and software: small size, middle size, standard platform league,
humanoid, 2D and 3D simulation \citep{KA+97}.
In the soccer simulation leagues \citep{ADL14}, the emphasis is on multi-robot team
work with partial and noisy information, in real-time. Each of the robots is
controlled by a separate program that receives sensor information from the
simulator as an input, and, asynchronously to sensor input, also decides on a
next action that is sent back to the simulator, several times a second. This
complexity of the environment, with continuous state and action spaces, together
with the opportunity to compete against each other, makes RoboCup soccer an
interesting testbed for multi-robot learning among many other applications.

To assist automated learning of team behavior, we provide a large dataset generated using 10 of the top participants in RoboCup 2016 or 2017. While it is possible to use the simulator for robot learning, we also generate additional data that is not normally available from playing other teams directly: We modified the simulator to record data from each robots local perspective, that is, with  the restricted views that depend on each robots situation and actions, and also include the sensor noise. 
In addition, for every step of the game (100\,ms), we recorded ground truth
information (such as positions and velocities) of all objects on the soccer field,
as well as basic actions of each robot. This ground truth information is usually
recorded in a logfile, but not available to teams during a match. 

To create our dataset, we ran all pairings of the selected teams with 25
repetitions of each game, that is, 1125 games in total. With 11 robots in each
team, a single game dataset consists of 22 local views plus a global
(ground-truth) view. These views are made available as CSV files 
(comma-separated values). We also provide the original logfiles that include
additional sensors, actions of each robot recorded as text files, and the
software we created to patch the simulator and to convert recordings into CSV.

The provided data are useful for various different tasks including imitation learning~\citep[e.g.,][]{BVE+13}, learning or testing of self-localization~\citep[e.g.,][]{Ols00}, predictive modeling of behavior, transfer learning and reinforcement learning~\citep[e.g.,][]{TS09}, and representation learning for time series data~\citep{MO+18}.
The next sections describe the environment, robots, and data in more detail. 

\section{Description of the environment}

The RoboCup Soccer Simulation Server \texttt{rcssserver} \citep{NM+98} is the software used for the
annual RoboCup competitions that have been held since 1997. It is hosted at
\URL{github.com/rcsoccersim/}. We
used the rcssserver version 15.3.0 to create the data for this paper. The
simulator implements the (2D) physics and rules of the game, and also handles
the interface to the programs controlling each player. By default, players use a
90 degree field of view, and receive visual information every 150\,ms. Throughout
the game, this frequency can actively be changed by each player individually to
75\,ms, 150\,ms, or 300\,ms, by changing the field of view to 45 degrees, 90 degrees,
or 180 degrees, respectively. Visual information is transmitted in the form of
(Lisp-like) lists of identified objects, with the level of detail of information
depending on object distances. Potential objects include all other players on
the field, the ball, and landmarks like goal posts, flags, and side lines. Each
player also receives additional status information, including energy levels,
referee decisions, and the state of the game, every 100\,ms. Each robot can issue
parameterized actions every 100\,ms, to control its locomotion, the direction of
its view, and its field of view. 
A more detailed description of the information transmitted can be found in the simulator manual~\citep{CDF+03}.

\section{Overview on the provided data}

In robotics, data collections often comprise lidar data recorded by laser scans.
This is very useful in many applications, e.g., field robotics, simultaneous
localization and mapping (SLAM) \citep{tong2013}, or specific purposes such as
agriculture \citep{chebrolu2017ijrr}. However, in many contexts, there is not
only one but several robots which may be observed. The data from RoboCup 
that we consider here includes information about other robots in the
environment and hence about the whole multi-robot system.

Data from multi-agent systems like the RoboCup or the multi-player real-time
strategy video game StarCraft \citep{LG+17} provide information on (simulated)
environments as in robotics. However, in addition, they contain data on other
agents and thus lay the basis for machine learning research to analyze and
predict agent behavior and strategies, which is important in applications such
as service robotics and multi-robot systems in general. To provide a diverse dataset, we include several teams from the last
two RoboCup competitions, allowing for different behaviors and strategies.


Perception and behavior of each robot during a game depends on the behavior of all other robots on the field. Game logfiles, that is, files containing ground truth information obtained from recording games, can be produced from the simulator, and are recorded in a binary format. Access to individual player percepts, however, is only possible from within the player code. To learn from behavior of other teams, it is useful to use the exact information that individual players receive, rather than the global (and noise-free) information in recorded logfiles. We therefore modified the simulator to additionally also record all local and noisy information as received by the robots on the field, in individual files for each player. This information is stored in the same format as it is sent to players. We also provide code to translate these individual logs into CSV files that contain relative positions and velocities.

We chose the top-ten teams from the RoboCup 2D soccer simulation world
championships 2016 in Leipzig (Germany) and 2017 in Nagoya (Japan) (see Figure~\ref{teams}). Team
binaries including further descriptions can be downloaded from:
\URL{archive.robocup.info/Soccer/Simulation/2D}. We played each team
against each other team for 25~times, resulting in about 17\,GB of data (zipped
CSV files). For each game, we record the original logfiles 
including message logs. We also generate files with ground truth data as well as
local player data in human-readable format. Finally, we made our generating
scripts available, so that they can be used to reproduce our results or to produce additional datasets using other
robotic soccer teams. There is also a smaller subset of 10~games where the
top-five teams play against each other once (163\,MB CSV files plus 217\,MB
original logfiles). Our data is available at
\URL{bitbucket.org/oliverobst/robocupsimdata/}.

\begin{figure}
\begin{center}\begin{tabular}{ll}
2016 & 2017\\ \hline
CSU\_Yunlu & CYRUS\\
Gliders & Fra-UNIted\\
HELIOS & HELIOS\\
Ri-one & HfutEngine\\
& MT \\
& Oxsy
\end{tabular}\end{center}
\caption{Teams from the RoboCup 2D soccer simulation world championships 2016 in
Leipzig (Germany) and 2017 in Nagoya (Japan) selected for the dataset.}
\label{teams}
\end{figure}

\section{Description of the ground truth data}

\begin{figure*}
\includegraphics[width=\textwidth]{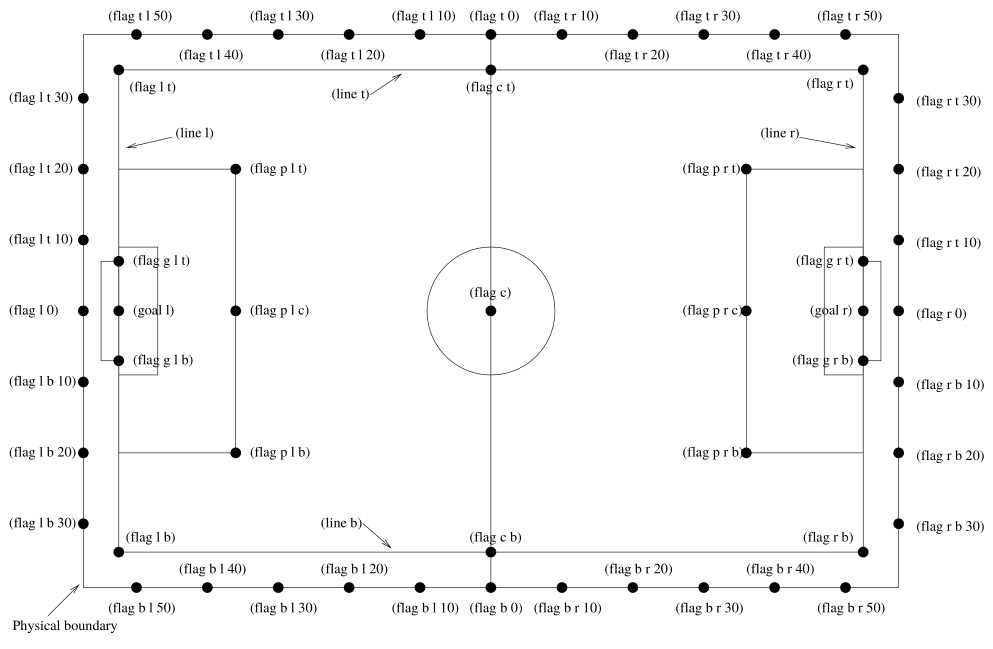}
\caption{Flags and lines in the robotic soccer simulation \citep[cf.][]{CDF+03}.}
\label{pitch}
\end{figure*}

According to rules of the world soccer association FIFA, a soccer pitch has the size of
$105\,\mathrm{m} \times 68\,\mathrm{m}$. This is adopted for the RoboCup soccer
simulation league. Nevertheless, the physical boundary of the area that may be
sensed by the robots has an overall size of $120\,\mathrm{m} \times
80\,\mathrm{m}$. For the localization of the robot players, the pitch is filled
with several landmark objects as depicted in Fig.~\ref{pitch}: flags~(f),
which are punctual objects, lines~(l), the goal~(g), and the penalty area~(p).
The origin of the coordinate system is the center point~(c). The pitch is
divided horizontally ($x$-direction) into a left~(l) and right~(r) half and
vertically ($y$-direction) into a top~(t) and a bottom~(b) half. Additional
numbers (10, 20, 30, 40, or 50) indicate the distance to the origin in meters.
Since every soccer game takes place on the same pitch, there is only one file
with infomation about the landmarks for all games that lists all these
coordinates, given as a table in CSV format, with name \verb'landmarks.csv'.
For example, the row \verb'f r t,52.5,34' says that the right top flag of the
pitch has the $(x,y)$-coordinates $(52.5,34)$. 

Further table files provide information about the respective game. The names of
all these files --  all naming conventions are summarized in Fig.~\ref{convention} -- contain the names of the
competing teams, the final scores for each team, possibly extended by the result
of a penalty shootout, a time stamp (when the game was recorded), and some
further identifier.

\begin{figure*}
\begin{enumerate}
  \item landmarks: \verb'landmarks.csv' (static information for all games, 1~file)
  \item game data: \verb'<time>-<team left>_<score left>-vs-<team right>_<score right>-<id>.csv'\\
	where \verb'<id>' =
	\begin{itemize}
	  \item \verb'parameters' (server configuration parameters, 1~file)
	  \item \verb'groundtruth' (logfile information for each game, 1~file)
	  \item \verb'<team name>_<player number>-<suffix>'\\
		where \verb'<suffix>' =
		\begin{itemize}
		  \item \verb'landmarks' (relative distances and angles to landmarks, $22$~files)
		  \item \verb'moving' (relative distances and angles to ball and other players, $22$~files)
		\end{itemize}
	\end{itemize} 
\end{enumerate}
\caption{Name conventions for the data files.}
\label{convention}
\end{figure*}

The central SoccerServer \citep{CDF+03} controls every virtual game with
built-in physics rules. When a game starts, the server may be configured by
several parameters which are collected in one file with the identifier
\verb'parameters'. For example, the row \verb'ball_decay,0.94' denotes that the
ball speed decreases by the specified factor \cite[cf.][]{SOM02b}. However, from
a robotics point of view, most of the information in this file is not very
relevant, like the stamina of the robots, the noise model, or the format of the coach
instructions. We thus skip further details here.

A soccer simulation game in the RoboCup 2D simulation league lasts 10\,mins in
total, divided into 6000~cycles where the length of each cycle is
100\,ms. Logfiles comprise information about the game,
in particular about the current positions of all players and the ball including
their velocity and orientation for each cycle. This information is collected for
the whole game in a table with the identifier \verb'groundtruth'. For each time
point, the play mode (e.g. kickoff), the current ball position coordinates and
its velocity is listed. Furthermore, the positions and velocities of each player
of the left (L) and the right (R) team including the goalkeeper (G) is stated.
For example, the column with head \verb'LG1 vx' contains the velocity of the
left goalkeeper in $x$-direction. Finally, information about the robots body and
head orientation and their view angle and quality is included. The absolute
 direction a player is facing (with respect to the pitch coordinate system) is
the sum of the body and head direction of that player.

\section{Description of local player data}

The visual sensor of the players reports the objects currently seen. The
information is automatically sent to players every sense step with a frequency
depending on the player view width and quality; by default it is set to 150\,ms. Thus, in
addition to the three files mentioned above, 44 more files are available for
each game. For each of the altogether 22 robots (ten field players and one
goalkeeper per team), two files with local player data is provided, hosting
information about where the respective player sees the landmarks and moving
objects, respectively.

The file with final identifier \verb'landmarks' provides the distances (in
meters) and angles (in degrees) to the respective landmarks relative to the
robot head orientation for each step. Analogously, the file with final
identifier \verb'moving' provides the actual relative distances and angles to
the ball and all other players. Sometimes the player number or even the team
name is not visible and hence unknown (u) to the robot. In this case, the respective piece of
information is left out. If data is not available at all, then this is marked by
\verb'NAN' in the respective table element. The server also provides information
about the velocity, stamina, (yellow and red) cards, and the commands (e.g. dash,
turn, or kick) of the robots. In some cases there is also information about the
observed state of other robots available, in particular, whether they are
kicking (k), tackling (t), or are a goalkeeper (g).

\section{Code}
The soccer simulator communicates with players using text messages in form of
lists, via UDP. As described above, the frequency of these messages is dependent
on the width of view that the player selects, one message every 150\,ms by
default. Every 100\,ms, players also receive status information, so-called body
messages. These individual messages are not recorded in the simulator logfiles.
Developers of teams can implement recording of messages that their own agents
receive, but in order to record messages sent to any team, the simulator
software had to be modified instead. Our code contains patches to the simulator
that allow recording of visual and body messages in individual files for each
player. The messages are stored in their original format to keep the amount of
additional processing during the game minimal. Playing a game with the modified
simulator will result in a number of recorded files:
\begin{itemize}
  \item the visual and body messages (two files for each player). The file names
	follow the same naming convention as the game data CSV files (cf.
	Fig.~\ref{convention}), but use the \verb'.rcv' suffix for visual
	messages and \verb'.rcb' for body messages.
  \item a recording of the game (the ground truth in a binary format), using the suffix \verb'.rcg'
  \item the commands from players as received by the simulator (in plain text), using \verb'.rcl' as a suffix 
\end{itemize}

To convert visual messages into CSV files, we provide a \texttt{see2csv.py}
Python program that translates player visual messages into two files: a CSV file
for moving objects (other players and the ball), and a CSV file with perceived
landmarks. 

To convert the simulator logfile (ground truth) into a CSV file, we provide
\texttt{rcg2csv}, a C++ program that is built using the open source
\texttt{librcsc} library (see~\URL{osdn.net/projects/rctools/releases/p3777}).
Logfiles are recorded at regular intervals of 100\,ms. \texttt{rcg2csv} can
optionally also store all simulation parameters in an additional CSV file. 

\section{Conclusions}

We released a large and unique dataset (RoboCupSimData), creating using a tournament
of top RoboCup simulation league teams. Creating this dataset required
modification of the simulation software. We ran 25 repetitions of 45 matches
that last 10\,mins, each. The dataset will allow a number of problems to be
investigated, from learning and testing approaches to self-localization,
predictive world-models, reinforcement learning. In this respect, the dataset
shall be useful for the whole robotics community.

\section*{Funding}

The research reported in this paper has been supported by the German Academic
Exchange Service (DAAD) by funds of the German Federal Ministry of Education and
Research (BMBF) in the Programmes for Project-Related Personal Exchange (PPP)
under grant no.~57319564 and Universities Australia (UA) in the
Australia-Germany Joint Research Cooperation Scheme within the project \emph{\underline{De}ep
\underline{Co}nceptors for Tempo\underline{r}al D\underline{at}a
M\underline{in}in\underline{g}} (Decorating).

\end{document}